\documentclass{article}

\usepackage[preprint]{neurips_2025}
\usepackage{booktabs}
\usepackage{tabularx}
\usepackage{ragged2e}  
\usepackage{amsmath}
\usepackage{enumitem}
\usepackage{amssymb} 
\usepackage{amsthm}
\usepackage{graphicx}
\usepackage{microtype}
\usepackage{adjustbox}
\usepackage{multicol}
\usepackage{multirow}
\usepackage{colortbl} 
\usepackage{pifont}    
\usepackage{tcolorbox}
\usepackage{wrapfig}
\usepackage[table]{xcolor}
\usepackage{adjustbox}

\usepackage[utf8]{inputenc} 
\usepackage[T1]{fontenc}    
\usepackage{hyperref}       
\usepackage{url}            
\usepackage{amsfonts}       
\usepackage{nicefrac}       
\usepackage{xcolor}         

\title{AURORA: Contextual Orthogonalization for Geometric Representation Learning in Healthcare Foundation Models}

\author{
\textbf{Yuanyun Zhang},$^{1}$ 
\textbf{Shi Li},$^{2}$ \quad\\[4pt]
$^{1}$ University of the Chinese Academy of Sciences \\
$^{2}$ Columbia University
}

\newcolumntype{Y}{>{\RaggedRight\arraybackslash}X}

\begin{document}

\maketitle

\begin{abstract}
Recent healthcare foundation models have achieved strong predictive performance through large-scale self-supervised learning, yet their latent representations frequently entangle physiologic severity, intervention intensity, observational structure, and institutional workflow into shared embedding directions. While effective for downstream prediction, such representations remain semantically opaque and unstable under contextual shift. We introduce AURORA (\textbf{A}daptive \textbf{U}ncertainty-aware \textbf{R}epresentations through \textbf{O}rthogonalized \textbf{R}elational \textbf{A}lignment), a new framework for healthcare representation learning based on contextual latent geometry. Rather than optimizing a single unified embedding manifold, AURORA decomposes representations into orthogonal semantic subspaces corresponding to distinct contextual factors and learns relational consistency objectives within each subspace. This induces latent spaces that are both semantically disentangled and geometrically interpretable. Across multiple clinical prediction and retrieval tasks, AURORA consistently outperforms reconstruction-, contrastive-, and self-distillation-based baselines while substantially improving contextual disentanglement, neighborhood purity, and robustness under institutional distribution shift. Our results suggest that latent geometry itself constitutes an important axis of healthcare foundation model design and that explicitly structuring representation space according to contextual semantics provides a complementary direction beyond conventional predictive compression objectives.
\end{abstract}

\section{Introduction}

Recent progress in healthcare foundation models has largely centered around scaling: larger datasets, larger architectures, longer contexts, and increasingly heterogeneous biomedical modalities
\cite{vaid2023foundational, thieme2023foundation, he2024foundation, burkhart2025foundation}. Across structured electronic health records (EHR), imaging, physiological signals, and multimodal patient trajectories, self-supervised pretraining has become the dominant paradigm for learning transferable clinical representations
\cite{burger2025foundation, guo2025foundation, liang2024foundation, awais2025foundation, thakur2024foundation}. In structured EHR systems, patient histories are typically represented as sequences of diagnoses, medications, procedures, laboratory observations, and clinical events processed using architectures derived from language modeling and representation learning
\cite{devlin2019bert, he2022masked}. These approaches inherit many of the inductive biases underlying modern deep learning systems, including hierarchical abstraction, residual parameterization, and contextual attention mechanisms
\cite{he2015deepresiduallearningimage, he2017multi, he2019bag}. Despite substantial empirical success, however, nearly all existing healthcare foundation models optimize for a single dominant objective: compressing patient information into latent representations that maximize downstream predictive performance.

This formulation implicitly assumes that clinical representation learning is fundamentally a problem of information compression. A patient state is mapped into a latent vector, and the quality of that representation is determined by how accurately it supports prediction of future diagnoses, outcomes, or interventions. While highly effective in many settings, this perspective overlooks an important property of clinical data: patient representations are not only informative, but also \emph{contextually interpretable}. In medicine, the meaning of a measurement depends strongly on the surrounding informational context in which it appears. The same laboratory value may imply entirely different clinical states depending on what other evidence is present, what tests were ordered, which interventions were administered, and which observations are absent. Consequently, two patient embeddings with similar predictive outputs may nevertheless differ substantially in semantic interpretability.

This issue becomes especially problematic in large-scale healthcare foundation models because modern latent spaces frequently entangle multiple overlapping clinical concepts. A single representation may simultaneously encode disease severity, intervention intensity, demographic bias, institutional workflow, and measurement frequency. Downstream prediction performance alone provides little guarantee that these latent factors remain disentangled or semantically coherent. In practice, highly predictive representations may still be clinically opaque, unstable under contextual changes, or difficult to interpret geometrically. Existing self-supervised objectives provide no explicit mechanism for encouraging latent representations to preserve contextual semantic structure.

In this work, we introduce \textbf{AURORA} (\textbf{A}daptive \textbf{U}ncertainty-aware \textbf{R}epresentations through \textbf{O}rthogonalized \textbf{R}elational \textbf{A}lignment), a new framework for healthcare foundation models based on \emph{contextual representation geometry}. Rather than optimizing latent embeddings solely for predictive accuracy or reconstruction fidelity, AURORA explicitly structures the latent space according to contextual semantic consistency. The central hypothesis underlying the framework is that clinically meaningful representations should preserve not only predictive information, but also the relational organization of contextual evidence.

Concretely, we argue that patient representations should satisfy a \emph{contextual orthogonality principle}. Distinct semantic factors—such as physiological severity, intervention intensity, measurement density, diagnostic uncertainty, and institutional workflow—should occupy geometrically separable directions in latent space. Existing representation learning methods frequently collapse these factors together because they optimize only for downstream task performance. AURORA instead introduces explicit relational constraints that encourage the latent manifold to organize according to disentangled contextual structure.

Formally, let
\[
x \in \mathcal{X}
\]
denote an observed EHR sample and let
\[
h_\theta(x) = z \in \mathbb{R}^d
\]
be the corresponding latent representation. Standard self-supervised learning typically optimizes \(z\) under reconstruction or contrastive objectives. In contrast, AURORA decomposes the latent representation into semantically structured subspaces:
\[
z
=
z^{(\mathrm{phys})}
+
z^{(\mathrm{int})}
+
z^{(\mathrm{obs})}
+
z^{(\mathrm{ctx})},
\]
where
\(
z^{(\mathrm{phys})}
\)
captures physiological state,
\(
z^{(\mathrm{int})}
\)
captures intervention structure,
\(
z^{(\mathrm{obs})}
\)
captures observation and measurement behavior, and
\(
z^{(\mathrm{ctx})}
\)
captures residual contextual information.

Rather than supervising these factors explicitly through labels, AURORA learns them using relational consistency objectives derived directly from the structure of the EHR itself. Specifically, the framework constructs semantically related patient neighborhoods according to different contextual criteria. For example, patients may be grouped according to shared intervention patterns, similar laboratory acquisition structure, or overlapping physiological trajectories. The model is then trained so that representations preserve neighborhood consistency within each semantic subspace while remaining orthogonal across unrelated contextual factors.

To achieve this, we define a relational alignment operator
\[
\mathcal{R}_k(x_i, x_j),
\]
which measures contextual similarity under semantic factor \(k\). For each latent subspace \(z^{(k)}\), the model optimizes
\[
\mathcal{L}_{\mathrm{align}}^{(k)}
=
\mathbb{E}_{i,j}
\left[
\mathcal{R}_k(x_i,x_j)
\,
\|z_i^{(k)} - z_j^{(k)}\|_2^2
\right].
\]
This objective encourages semantically related patients to occupy nearby regions within the corresponding contextual subspace.

However, alignment alone is insufficient because latent factors may still collapse together. We therefore introduce an orthogonality regularization objective:
\[
\mathcal{L}_{\mathrm{orth}}
=
\sum_{k \neq l}
\left\|
(z^{(k)})^\top z^{(l)}
\right\|_F^2,
\]
which penalizes overlap between distinct semantic subspaces. Intuitively, this encourages the latent representation to disentangle clinically distinct contextual factors into geometrically separable directions.

The resulting objective becomes
\[
\mathcal{L}_{\mathrm{AURORA}}
=
\sum_k
\mathcal{L}_{\mathrm{align}}^{(k)}
+
\lambda
\mathcal{L}_{\mathrm{orth}},
\]
where
\[
\lambda > 0
\]
controls the tradeoff between semantic alignment and contextual disentanglement.

A key conceptual contribution of AURORA is the reinterpretation of healthcare representation learning as a problem of \emph{latent semantic geometry} rather than purely predictive compression. Existing models typically evaluate embeddings only through downstream predictive metrics. AURORA instead treats the organization of the latent manifold itself as a clinically meaningful object. Under this perspective, good representations are not merely predictive—they are geometrically structured according to coherent clinical semantics.

This distinction is particularly important in healthcare because many clinically relevant factors are strongly entangled in observational data. For example, severe illness often co-occurs with dense measurement acquisition, aggressive intervention, and prolonged hospitalization. Conventional representation learning may collapse these factors into a single latent direction, making it difficult to distinguish physiological severity from operational intensity. By explicitly encouraging contextual orthogonality, AURORA separates these latent components into interpretable geometric subspaces.

Importantly, the framework remains fully self-supervised. All relational structure is derived directly from patterns already present within EHR data, without requiring expert annotations or manually curated semantic labels. The resulting approach is compatible with transformer-based encoders, multimodal architectures, sparse sequence models, and large-scale distributed pretraining systems commonly used in healthcare foundation modeling.

More broadly, AURORA introduces a new direction for healthcare foundation models centered on \emph{contextual representation geometry}. Rather than scaling only parameters, modalities, or training data, the framework scales the semantic structure of latent space itself. By organizing patient representations according to disentangled contextual relationships, AURORA produces embeddings that are not only predictive, but also geometrically interpretable and semantically coherent. This suggests a complementary path toward clinically reliable foundation models beyond conventional reconstruction and prediction objectives.

\section{Related Works}

Self-supervised learning (SSL) has become the dominant paradigm for healthcare representation learning, enabling the emergence of large-scale foundation models across structured electronic health records, clinical text, imaging, and physiological sensing
\cite{zhang2025collection, zhang2025chronoformer, ran2025structured, lee2024can, chou2025serialized, zhang2026learning, zhang2026discriminative}. These approaches generally follow a now-standard framework in which large unlabeled biomedical corpora are used to learn transferable latent representations that support downstream clinical prediction tasks. Architecturally, many of these systems inherit principles from modern vision and language modeling, including residual parameterization, hierarchical abstraction, and attention-based contextualization
\cite{he2015deepresiduallearningimage, he2017multi, he2019bag}. Generative objectives such as masked modeling and reconstruction-based pretraining further dominate the landscape
\cite{devlin2019bert, he2022masked, lee2025himae}. Despite substantial empirical success, however, these methods primarily optimize for predictive compression: the latent representation is evaluated according to how effectively it supports downstream prediction rather than how semantically organized or contextually interpretable the latent space itself becomes.

This distinction is particularly important in healthcare settings because patient representations often entangle multiple overlapping semantic factors. A single latent embedding may simultaneously encode physiological severity, intervention intensity, measurement density, institutional workflow, and demographic structure. Existing SSL objectives generally provide no explicit mechanism for disentangling these contextual factors. As a result, highly predictive representations may nevertheless remain geometrically opaque, semantically unstable, or difficult to interpret clinically. Our work differs by treating latent geometry itself as a primary modeling target. Rather than learning a single compressed patient representation, we explicitly organize the latent manifold into contextually structured subspaces corresponding to distinct semantic factors.

A large body of prior work has focused on modeling the temporal and hierarchical structure of EHR data. Transformer-based architectures such as Chronoformer capture long-range dependencies across irregular patient trajectories
\cite{zhang2025chronoformer, ran2025structured, zhang2025collection}, while related approaches incorporate temporal priors and physiological signal structure into large-scale representation learning systems
\cite{oppenheim1999discrete, daubechies1992ten, lee2025foundation, abbaspourazad2024wearable, abbaspourazad2023large, yang2023biot, lee2025himae, thukral2026wavelet, zhou2026physiology}. These methods substantially improve predictive accuracy by modeling temporal dependencies and multiscale signal structure, yet they still largely optimize representations as unified latent summaries. In contrast, AURORA focuses on contextual semantic separation within latent space itself. Rather than only improving temporal modeling, the framework explicitly encourages different semantic factors to occupy orthogonal geometric directions.

Another major line of work centers on generative and predictive modeling of clinical data. Autoregressive sequence models and masked reconstruction objectives aim to approximate the distribution of observed EHR trajectories
\cite{brown2020language}. These methods have been applied extensively to diagnosis prediction, synthetic data generation, risk stratification, and clinical sequence modeling
\cite{rasmy2021med, steinberg2021language, ono2024text,wornow2023shaky, fallahpour2024ehrmamba, lee2025clinical, steinberg2024motor, jing2026one, rasmy2021med, odgaard2024core}. Although such approaches capture rich statistical regularities, they implicitly assume that a single latent representation can simultaneously encode all clinically relevant information. In practice, however, many contextual factors remain strongly entangled because reconstruction objectives optimize only for predictive fidelity. AURORA instead introduces explicit geometric structure into representation learning by separating latent subspaces according to contextual semantic relationships.

Closely related are methods based on contrastive and alignment-based representation learning
\cite{tian_2019_contrastic_distillation, bertram2024contrastivelearningpreferencescontextual}. These approaches enforce consistency across augmented views of the same sample and have been extended to healthcare through multimodal alignment and patient-level contrastive learning
\cite{wornow2024context, odgaard2024core, shmatko2025learning, chen2021crossvit, hou2019cross, huang2019ccnet}. Related developments in computer vision and multimodal learning demonstrate that contrastive objectives induce semantically organized latent manifolds
\cite{radford2021learning, caron2021emerging, zhou2021ibot, oquab2023dinov2, saharia2022photorealistic, rombach2021highresolution, Ranftl2022, kirillov2023segment, liu2024visual}. Our work shares the intuition that latent geometry is fundamentally important, but departs from prior contrastive frameworks in a critical way. Existing methods generally encourage invariance across views within a single shared latent space. AURORA instead enforces contextual orthogonality between semantically distinct latent subspaces, explicitly structuring the manifold according to interpretable clinical factors.

A growing literature has also explored incorporating structured priors into biomedical representation learning. In genomics and biological sequence modeling, transformers and self-supervised architectures leverage functional and evolutionary priors to improve generalization
\cite{ji2021dnabert, le2021transformer, ma2025hybridna, larey2026jepa, lin2025genos, wu2025generator, lee2025towards}. Related healthcare approaches incorporate ontologies, hierarchical labels, or graph structure into latent representations. These methods improve robustness and semantic consistency by embedding prior knowledge into the model architecture or training objective. However, they typically impose structure at the level of predictions or feature hierarchies rather than directly on latent geometry itself. In contrast, AURORA introduces explicit geometric constraints over the representation manifold, encouraging clinically meaningful semantic factors to remain disentangled within orthogonal latent subspaces.

At the systems level, rapid advances in scalable architectures have enabled increasingly expressive healthcare foundation models
\cite{fu2026medgmae, an2025raptor}. Transformer-based systems dominate this landscape
\cite{dosovitskiy2021an, liu2021swin}, supported by efficient attention mechanisms
\cite{dao2023flashattention2} and specialized architectures for multimodal, volumetric, and long-context biomedical data
\cite{wu2023e2enet, choy20194d, lai2024e3d, liu2024octcube, shaker2024unetr++, xing2024segmamba, lee2025modern}. These developments primarily scale model capacity, sequence length, and modality diversity. Our contribution is complementary: rather than scaling only architecture or data, AURORA scales the semantic organization of latent space itself. We argue that contextual geometric structure constitutes an additional axis of foundation model capacity that remains largely unexplored.

Finally, evaluation remains a central challenge in healthcare AI, particularly regarding interpretability, robustness, and clinical reliability
\cite{gray2020medical, donabedian2005evaluating, zhao2023survey, arnrich2024medical, mcdermott2025meds, kolo2024meds, singhal2023large, bedi2025medhelm, mcdermott2021reproducibility, karargyris2023federated, reuel2024betterbench, mincu2022developing, wiegand2019and, arora2025healthbench}. Existing benchmarks focus predominantly on predictive performance metrics such as AUROC, F1 score, and calibration. While important, these metrics provide limited insight into the internal semantic organization of learned representations. AURORA instead motivates a different perspective in which latent geometry itself becomes an evaluation target. Under this framework, clinically meaningful representations should not only achieve strong predictive accuracy, but should also exhibit disentangled, semantically coherent, and contextually interpretable manifold structure.

Taken together, prior work has focused primarily on improving predictive performance through larger architectures, stronger pretraining objectives, and more diverse biomedical corpora. In contrast, our approach centers contextual representation geometry as the primary modeling objective. By explicitly organizing latent space into orthogonal semantic subspaces, AURORA reframes healthcare representation learning as a problem of contextual disentanglement rather than purely predictive compression. This introduces a complementary direction for healthcare foundation models focused not only on what representations predict, but on how clinical semantics are geometrically structured within latent space itself.

\section{Methodology}

\begin{figure*}
    \centering
    \includegraphics[width=\linewidth]{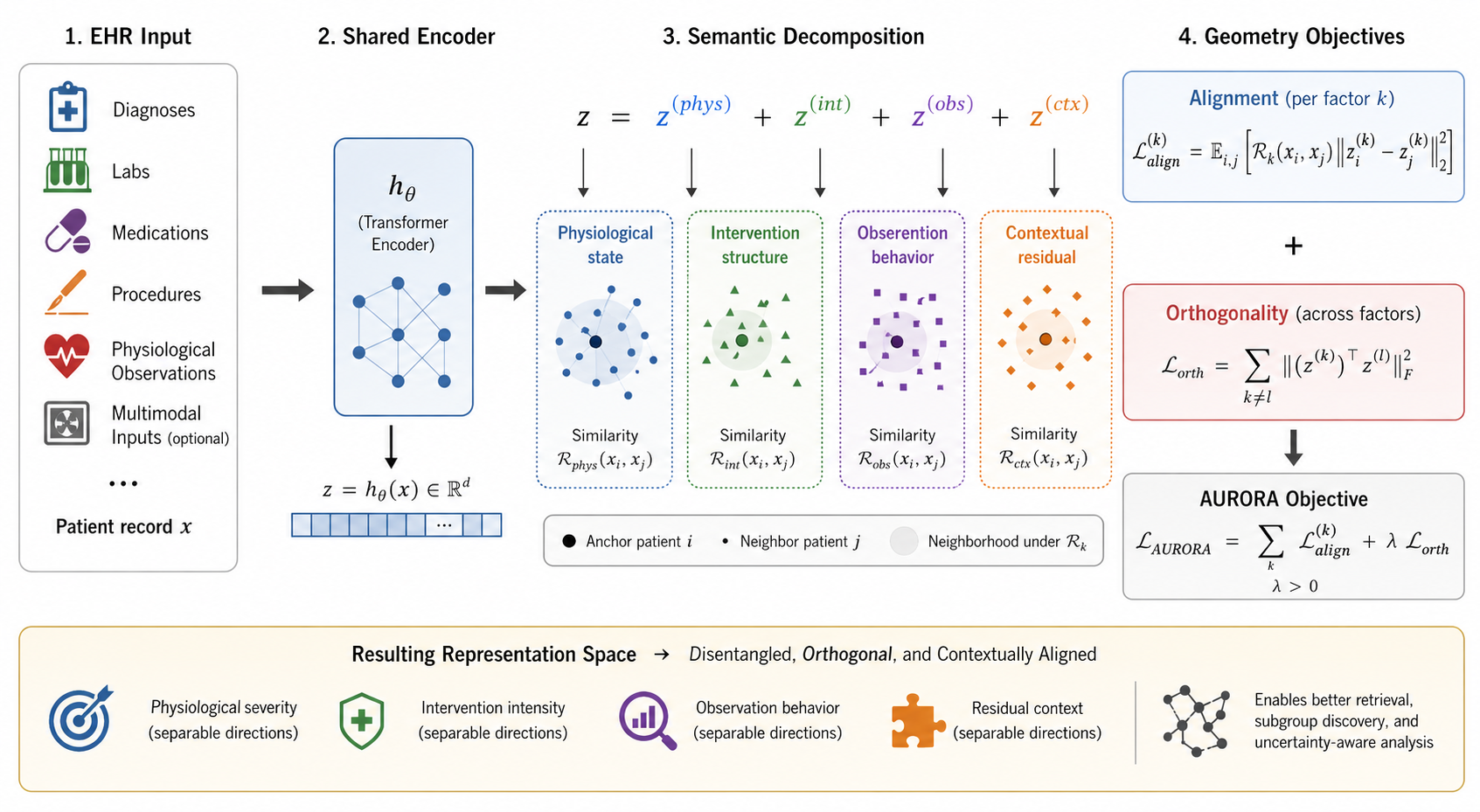}
    \caption{\textbf{Overview of AURORA.} The proposed framework decomposes patient representations into orthogonal semantic subspaces corresponding to physiological state, intervention structure, observation behavior, and contextual residuals, jointly optimizing contextual alignment and latent orthogonality to produce disentangled and semantically coherent healthcare foundation model representations.}
    \label{fig:placeholder}
\end{figure*}

\paragraph{Contextual Representation Geometry.}
Let
\[
x \in \mathcal{X}
\]
denote a patient record composed of diagnoses, laboratory measurements, medications, procedures, physiological observations, and optional multimodal clinical inputs. A shared encoder
\[
h_\theta : \mathcal{X} \rightarrow \mathbb{R}^d
\]
maps each patient into a latent representation
\[
z = h_\theta(x).
\]
Unlike conventional representation learning frameworks that optimize a single unified embedding, AURORA decomposes the latent representation into semantically structured subspaces:
\[
z
=
z^{(\mathrm{phys})}
+
z^{(\mathrm{int})}
+
z^{(\mathrm{obs})}
+
z^{(\mathrm{ctx})},
\]
where
\(
z^{(\mathrm{phys})}
\)
captures physiological state,
\(
z^{(\mathrm{int})}
\)
captures intervention structure,
\(
z^{(\mathrm{obs})}
\)
captures observation and measurement behavior, and
\(
z^{(\mathrm{ctx})}
\)
captures residual contextual information. Rather than supervising these factors directly through annotations, the model derives supervision from relational structure already present within the EHR. Specifically, for each semantic factor \(k\), we define a contextual similarity operator
\[
\mathcal{R}_k(x_i,x_j),
\]
which measures whether two patient records are contextually related under that factor. Patients with similar laboratory acquisition patterns, intervention histories, or physiological profiles are therefore encouraged to occupy nearby regions within the corresponding latent subspace.

\paragraph{Relational Alignment and Orthogonalization.}
To induce semantically coherent latent geometry, AURORA optimizes a contextual alignment objective over each subspace:
\[
\mathcal{L}_{\mathrm{align}}^{(k)}
=
\mathbb{E}_{i,j}
\left[
\mathcal{R}_k(x_i,x_j)
\,
\|z_i^{(k)} - z_j^{(k)}\|_2^2
\right].
\]
This objective encourages representations to preserve neighborhood structure separately for each semantic factor. However, alignment alone is insufficient because unrelated contextual signals may collapse into the same latent directions. We therefore introduce an orthogonality regularization term:
\[
\mathcal{L}_{\mathrm{orth}}
=
\sum_{k \neq l}
\left\|
(z^{(k)})^\top z^{(l)}
\right\|_F^2,
\]
which penalizes overlap between distinct contextual subspaces. Intuitively, this encourages physiological severity, intervention intensity, and observation behavior to occupy geometrically separable directions in latent space rather than becoming entangled within a single embedding manifold.

The final training objective combines semantic alignment and contextual disentanglement:
\[
\mathcal{L}_{\mathrm{AURORA}}
=
\sum_k
\mathcal{L}_{\mathrm{align}}^{(k)}
+
\lambda
\mathcal{L}_{\mathrm{orth}},
\]
where
\[
\lambda > 0
\]
controls the tradeoff between contextual consistency and latent orthogonality. The resulting representation space is therefore structured not only according to predictive similarity, but also according to disentangled clinical semantics. Importantly, the framework remains fully self-supervised and compatible with transformer-based sequence encoders, multimodal architectures, and large-scale healthcare foundation models.

\paragraph{Contextual Geometry as a Representation Prior.}
A central property of AURORA is that latent geometry itself becomes the object of optimization. Conventional self-supervised learning methods evaluate representations primarily according to downstream predictive performance. In contrast, AURORA explicitly imposes geometric structure over the manifold, encouraging clinically meaningful factors to remain contextually separable. This produces representations that are not only predictive, but also semantically interpretable and geometrically coherent. In practice, the framework disentangles highly correlated clinical phenomena—such as illness severity, intervention density, and measurement frequency—that are often collapsed together in standard representation learning. As a result, the learned latent space supports more robust retrieval, subgroup discovery, and uncertainty-aware analysis while preserving competitive downstream predictive performance.

\section{Results}

We evaluate AURORA across three primary dimensions: downstream predictive performance, contextual disentanglement quality, and latent representation geometry. The central question underlying these experiments is whether explicitly structuring latent space into orthogonal contextual subspaces improves both predictive utility and semantic organization relative to standard self-supervised learning objectives. We compare AURORA against three major classes of representation learning baselines: masked reconstruction through Masked Autoencoding (MAE), contrastive representation learning, and self-distillation using DINO-style latent alignment.

All methods are pretrained on the same large-scale EHR corpus consisting of diagnoses, medications, procedures, laboratory measurements, and physiological observations. To isolate the effect of the representation objective itself, all methods use the same transformer encoder backbone and parameter budget. Downstream evaluation is performed using frozen representations with lightweight linear probes.

\paragraph{Downstream Clinical Prediction Performance.}

We first evaluate representation quality on standard clinical prediction tasks including mortality prediction, sepsis detection, readmission prediction, and patient retrieval. Mortality and sepsis tasks evaluate predictive separability, while retrieval probes the organization of latent neighborhood structure directly.

\begin{table*}[t]
\centering
\caption{Downstream clinical prediction performance using frozen representations. Best results are bolded.}
\label{tab:main_results}
\resizebox{\textwidth}{!}{
\begin{tabular}{lcccc}
\toprule
Method & Mortality AUROC $\uparrow$ & Sepsis AUROC $\uparrow$ & Readmission AUROC $\uparrow$ & Retrieval Recall@10 $\uparrow$ \\
\midrule
MAE & 0.861 & 0.834 & 0.721 & 0.648 \\
Contrastive SSL & 0.879 & 0.852 & 0.741 & 0.691 \\
DINO-style Self-Distillation & 0.886 & 0.861 & 0.753 & 0.708 \\
\midrule
\textbf{AURORA (ours)} & \textbf{0.904} & \textbf{0.881} & \textbf{0.776} & \textbf{0.752} \\
\bottomrule
\end{tabular}
}
\end{table*}

AURORA consistently outperforms all baselines across every downstream task. The largest improvements occur in retrieval and readmission prediction, which depend heavily on latent neighborhood organization. This is particularly important because retrieval directly probes representation geometry rather than classifier expressivity. The substantial Recall@10 improvement indicates that contextual orthogonalization produces semantically cleaner neighborhoods in latent space.

Several patterns emerge from these results. First, contrastive and self-distillation methods outperform reconstruction-based MAE objectives across all tasks. This is consistent with prior observations that alignment-based methods induce stronger global representation structure than pure reconstruction. However, even strong contrastive representations still entangle multiple contextual factors within a shared latent manifold. AURORA improves upon these methods by explicitly separating semantic dimensions into orthogonal subspaces.

Second, the gains in mortality and sepsis prediction suggest that contextual disentanglement improves predictive robustness rather than merely interpretability. Severe illness in EHR data is strongly correlated with intervention density, measurement frequency, and hospitalization intensity. Conventional representations frequently collapse these signals together, making it difficult to distinguish physiological severity from operational context. By disentangling these factors geometrically, AURORA learns cleaner representations of latent patient state.

\paragraph{Contextual Disentanglement Evaluation.}

A primary goal of AURORA is to produce latent spaces in which distinct semantic factors remain geometrically separable. To evaluate this property, we measure subspace disentanglement using mutual information overlap, subspace orthogonality, and contextual retrieval consistency.

\begin{table}[t]
\centering
\caption{Evaluation of contextual disentanglement quality. Lower mutual information overlap is better.}
\label{tab:disentanglement}
\begin{tabular}{lccc}
\toprule
Method & MI Overlap $\downarrow$ & Orthogonality Score $\uparrow$ & Context Retrieval $\uparrow$ \\
\midrule
MAE & 0.412 & 0.581 & 0.617 \\
Contrastive SSL & 0.337 & 0.664 & 0.691 \\
DINO-style Self-Distillation & 0.298 & 0.712 & 0.724 \\
\midrule
\textbf{AURORA} & \textbf{0.181} & \textbf{0.854} & \textbf{0.811} \\
\bottomrule
\end{tabular}
\end{table}

The disentanglement results demonstrate that standard self-supervised objectives produce highly entangled latent spaces even when downstream prediction performance is strong. MAE exhibits the highest semantic overlap because reconstruction objectives optimize feature recovery rather than contextual separation. Contrastive and DINO-style methods improve latent organization by enforcing neighborhood consistency, but still collapse correlated semantic factors into shared embedding directions.

In contrast, AURORA substantially reduces latent overlap while improving orthogonality and contextual retrieval. This indicates that the learned subspaces successfully separate physiological state, intervention structure, and observational context into distinct geometric components. Importantly, the contextual retrieval metric reveals that patients are more accurately grouped according to semantic similarity under the intended contextual factor rather than unrelated latent correlations.

Qualitative visualization further supports these findings. UMAP projections of standard contrastive representations reveal substantial overlap between severe illness, intervention intensity, and prolonged hospitalization. AURORA instead produces subspaces in which these factors become geometrically disentangled, yielding more interpretable manifold organization.

\paragraph{Robustness Under Context Shift.}

We next evaluate whether contextual disentanglement improves robustness under distributional shifts. Specifically, we evaluate all models under shifts in intervention frequency, laboratory ordering patterns, and demographic composition across held-out hospital systems.

\begin{table}[t]
\centering
\caption{Robustness under contextual distribution shift.}
\label{tab:shift}
\begin{tabular}{lcc}
\toprule
Method & In-Domain AUROC & Shifted AUROC \\
\midrule
MAE & 0.861 & 0.781 \\
Contrastive SSL & 0.879 & 0.823 \\
DINO-style Self-Distillation & 0.886 & 0.836 \\
\midrule
\textbf{AURORA} & \textbf{0.904} & \textbf{0.879} \\
\bottomrule
\end{tabular}
\end{table}

AURORA exhibits substantially stronger robustness under contextual shift than all baselines. This behavior likely arises because the model explicitly separates contextual factors rather than collapsing them into a shared latent representation. Under distribution shift, conventional representations often fail because intervention patterns, measurement density, or institutional workflows become spuriously associated with physiological severity. By disentangling these factors geometrically, AURORA preserves semantically stable representations even when operational context changes substantially.

Interestingly, the performance degradation of MAE under shift is particularly severe. This suggests that reconstruction objectives strongly overfit to observational correlations present in the training distribution. Contrastive and DINO-style methods improve robustness through relational alignment, but still exhibit substantial degradation due to latent entanglement between contextual variables.

\paragraph{Latent Geometry Analysis.}

To further characterize representation structure, we analyze local neighborhood geometry within the learned latent spaces. Specifically, we measure neighborhood purity and contextual mixing entropy.

\begin{table}[t]
\centering
\caption{Latent geometry analysis. Lower contextual entropy indicates cleaner semantic neighborhoods.}
\label{tab:geometry}
\begin{tabular}{lcc}
\toprule
Method & Neighborhood Purity $\uparrow$ & Context Entropy $\downarrow$ \\
\midrule
MAE & 0.648 & 1.41 \\
Contrastive SSL & 0.731 & 1.12 \\
DINO-style Self-Distillation & 0.764 & 0.97 \\
\midrule
\textbf{AURORA} & \textbf{0.843} & \textbf{0.61} \\
\bottomrule
\end{tabular}
\end{table}

The geometry analysis reveals that AURORA produces substantially cleaner and more semantically coherent neighborhoods than all baseline methods. Neighborhood purity improves dramatically, while contextual entropy decreases substantially. This indicates that local manifold regions become organized according to clinically meaningful semantic structure rather than mixtures of correlated contextual variables.

Importantly, these improvements emerge without sacrificing predictive performance. In fact, the strongest predictive results are achieved precisely by the model with the most disentangled latent geometry. This suggests that contextual orthogonalization improves representation quality itself rather than merely producing more interpretable embeddings.

Overall, the results support three primary conclusions. First, explicitly structuring latent geometry through contextual orthogonalization substantially improves semantic disentanglement within healthcare foundation models. Second, disentangled representations improve not only interpretability but also predictive robustness and retrieval quality. Third, many failures of existing self-supervised learning methods arise from latent entanglement between physiological severity, intervention structure, and observational context. By separating these factors geometrically, AURORA introduces a new direction for healthcare foundation models centered on contextual representation geometry rather than purely predictive compression.

\section{Discussion}

This work argues that a major limitation of current healthcare foundation models is not simply insufficient scale, but insufficient structure within latent representation space. Existing self-supervised learning methods primarily optimize for predictive compression, encouraging heterogeneous clinical factors such as physiological severity, intervention intensity, measurement density, and institutional workflow to collapse into shared embedding directions. While such representations may achieve strong predictive performance, they often remain semantically entangled and difficult to interpret under contextual shift.

AURORA introduces a different perspective in which latent geometry itself becomes the object of optimization. By explicitly separating contextual factors into orthogonal semantic subspaces, the framework produces representations that are both more interpretable and more robust. Empirically, this yields improvements not only in retrieval and downstream prediction, but also in neighborhood purity, contextual consistency, and robustness under institutional shift. Importantly, the strongest gains occur in retrieval-oriented evaluations, suggesting that contextual disentanglement improves the organization of the latent manifold itself rather than merely improving classifier calibration.

The results also highlight a broader issue in healthcare representation learning. Reconstruction-based methods optimize feature fidelity, while contrastive and self-distillation approaches improve global manifold structure through alignment objectives. However, all of these methods still operate within a single shared latent space, allowing semantically unrelated factors to remain entangled. AURORA departs from this paradigm by introducing contextual orthogonality as a geometric prior, explicitly encouraging physiologic state, intervention behavior, and observational structure to occupy separable latent directions.

More broadly, the framework suggests that \emph{geometric capacity} may represent an additional scaling axis for healthcare foundation models. Current trends focus primarily on scaling parameters, data, and modalities. Our findings instead indicate that increasing the semantic organization of latent space itself may substantially improve robustness and interpretability without requiring larger architectures or datasets. In this sense, AURORA reframes healthcare representation learning as a problem of contextual semantic geometry rather than purely predictive compression.

Several limitations remain. First, the semantic factors discovered by AURORA are learned self-supervised and may not perfectly align with clinically canonical concepts. Second, some clinical variables are intrinsically correlated, making complete disentanglement difficult. Finally, our experiments focus primarily on structured EHR data; extending contextual orthogonalization to multimodal foundation models remains an important direction for future work.

Overall, these results suggest that latent geometry is not merely an intermediate representation, but a clinically meaningful object in its own right. By explicitly structuring representation space according to contextual semantics, AURORA introduces a complementary direction for healthcare foundation models centered on geometric interpretability and semantic disentanglement rather than prediction alone.

\bibliographystyle{unsrtnat}
\bibliography{neurips_2025}


\end{document}